\documentclass[letterpaper]{article} 
\usepackage{aaai2026}  
\usepackage{times}  
\usepackage{helvet}  
\usepackage{courier}  
\usepackage[hyphens]{url}  
\usepackage{graphicx} 
\usepackage{natbib}  
\usepackage{caption} 
\usepackage{algorithm}
\usepackage{algorithmic}

\usepackage{xcolor}

\usepackage{newfloat}
\usepackage{listings}
\DeclareCaptionStyle{ruled}{labelfont=normalfont,labelsep=colon,strut=off} 
\floatstyle{ruled}
\newfloat{listing}{tb}{lst}{}
\floatname{listing}{Listing}
\usepackage{comment}
\usepackage{booktabs}
\usepackage{amsmath}
\usepackage{amsfonts}
\usepackage{multirow}
\usepackage{array}
\usepackage{pifont}     
\usepackage{xcolor}     
\usepackage{tipa}
\usepackage{tcolorbox}
\usepackage{spverbatim}
\usepackage{fancyvrb}
\usepackage{subcaption}

\definecolor{textemb}{HTML}{66c2a5}
\definecolor{llmemb}{HTML}{fc8d62}
\definecolor{multiemb}{HTML}{8da0cb}
\definecolor{vlmemb}{HTML}{e78ac3}

\definecolor{gptfive}{HTML}{14B8A6}
\definecolor{gptfivemini}{HTML}{5EEAD4}
\definecolor{geminipro}{HTML}{6366F1}
\definecolor{geminiflash}{HTML}{A5B4FC}
\definecolor{claude}{HTML}{F59E0B}
\definecolor{gptoss}{HTML}{F43F5E}

\definecolor{cS}{HTML}{4c78a8}
\definecolor{cP}{HTML}{72b7b2}
\definecolor{cTextEmb}{HTML}{5e4fa2}
\definecolor{cLLMEmb}{HTML}{8073ac}
\definecolor{cMultiEmb}{HTML}{b2abd2}
\definecolor{cVLMEmb}{HTML}{d8daeb}
\definecolor{cY}{HTML}{f58518}
\definecolor{cN}{HTML}{ffbf79}

\title{Memes-as-Replies: Can Models Select Humorous Manga Panel Responses?}
\author{
    Ryosuke Kohita,
    Seiichiro Yoshioka
}
\affiliations{
    CyberAgent\\

    \{kohita\_ryosuke, yoshioka\_seiichiro\}@cyberagent.co.jp
}

\usepackage{bibentry}

\nocopyright

\begin{document}

\maketitle

\begin{abstract}
Memes are a popular element of modern web communication, used not only as static artifacts but also as interactive replies within conversations. While computational research has focused on analyzing the intrinsic properties of memes, the dynamic and contextual use of memes to create humor remains an understudied area of web science.
To address this gap, we introduce the Meme Reply Selection task and present \textsc{MaMe-Re} (Manga Meme Reply Benchmark),\footnote{\url{https://will.be.published.soon}. Pronounced as \textipa{/meIm ri:/} (Maym-Re). } a benchmark of 100,000 human-annotated pairs (500,000 total annotations from 2,325 unique annotators) consisting of openly licensed Japanese manga panels and social media posts. Our analysis reveals three key insights: (1) large language models (LLMs) show preliminary evidence of capturing complex social cues such as exaggeration, moving beyond surface-level semantic matching; (2) the inclusion of visual information does not improve performance, revealing a gap between understanding visual content and effectively using it for contextual humor; (3) while LLMs can match human judgments in controlled settings, they struggle to distinguish subtle differences in wit among semantically similar candidates. These findings suggest that selecting contextually humorous replies remains an open challenge for current models.
\end{abstract}


\section{Introduction}
\label{sec:intro}
\begin{figure}[t]
  \centering
  \includegraphics[width=0.99\linewidth]{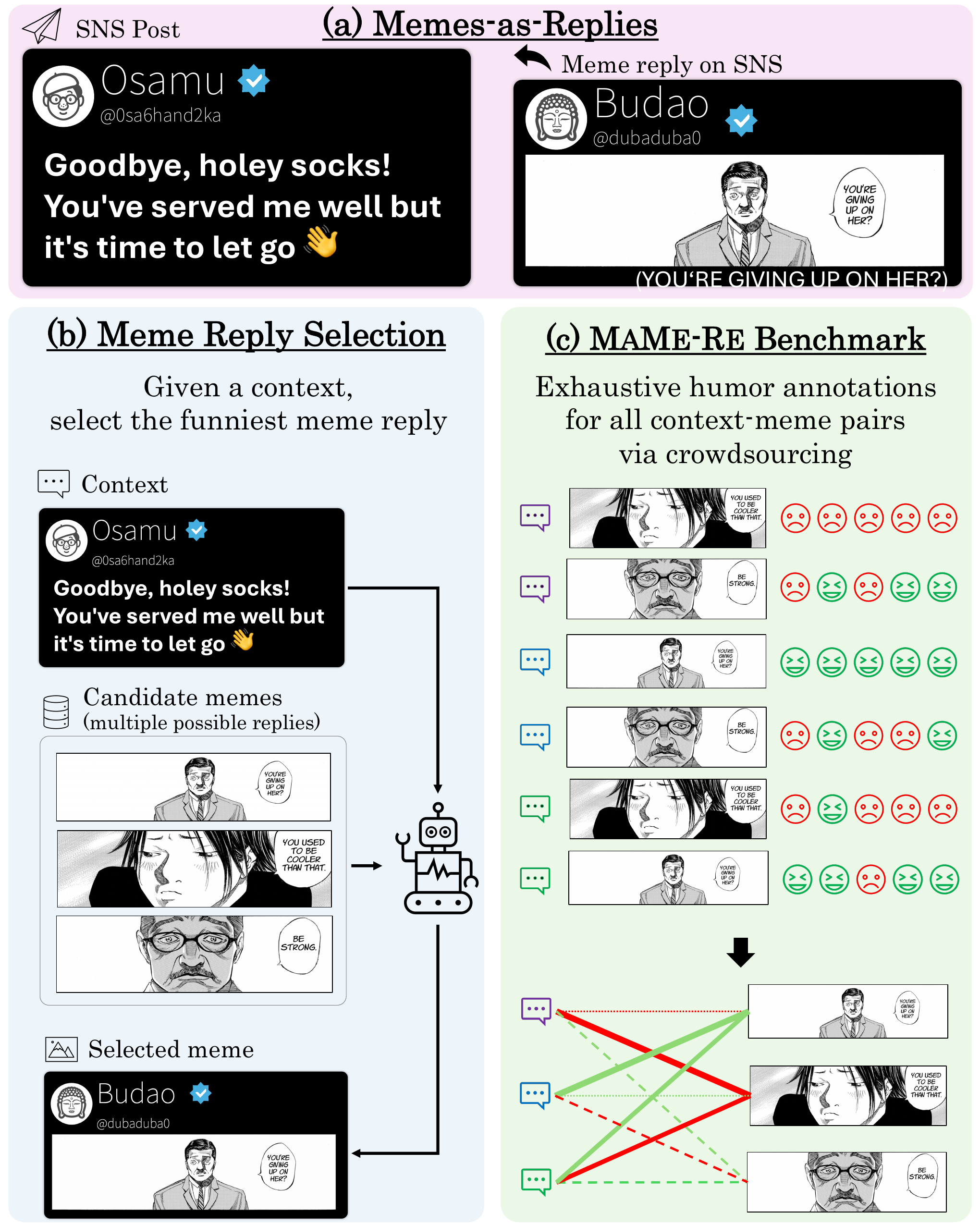}
  \caption{\small Overview of memes-as-replies. 
(a) Example of meme use on SNS. 
(b) Visualization of the Meme Reply Selection task. 
(c) \textsc{MaMe-Re} benchmark with crowdsourced humor labels.}
 \label{fig:overview}
\end{figure}

Memes are a popular element of modern web communication~\cite{joshi-2024-contextualizing}. People frequently use memes not just as static content, but as interactive replies within conversations~\cite{wang-2019-culturally, fei2021expressivecommunicationinternetmemes}. This practice enables creative and nuanced expressions that are difficult to convey through text alone~\cite{grundlingh2018memes}. For example, in Figure~\ref{fig:overview}(a), a user posts about their worn-out socks. In response, another user replies with a manga panel that dramatically asks, ``\textit{YOU'RE GIVING UP ON HER?}'' This reply creates humor by repurposing the manga panel to treat the socks as a person. Such meme replies can express a wide range of humor, from playful teasing to irony, all of which rely on a deep understanding of context and implicit cultural knowledge~\cite{davison2012language, yus2018identity, sharma-2022-ijicai}. While humans can intuitively grasp the social cues required for such humor, it reveals a significant limitation of current computational approaches~\cite{nguyen-ng-2024-computational}. This gap between widespread human practice and limited computational understanding motivates our work to systematically analyze this core web phenomenon.

Most computational research on memes treats them as static content, like an isolated photograph, for analyzing intrinsic properties for tasks like harmfulness detection or humorous captioning~\cite{shah-etal-2024-memeclip, sharma-etal-2020-semeval, wang-2024-memecraft, loakman2025whoslaughingnowoverview}. In contrast, our work focuses on the interactive function of memes as replies within a conversation. As illustrated in Figure~\ref{fig:overview}(b), we define the \textit{Meme Reply Selection} task, where the goal is to choose the funniest meme given a conversational context. This perspective recognizes that humor is not an intrinsic property of the meme itself, but an emergent quality of its interaction with context. This distinguishes our task from other forms of dialogue media selection, such as emojis or stickers, which typically prioritize simple semantic relevance~\cite{wang-jurgens-2021-animated-picture, lu-2023-towards-building}. We argue that modeling this dynamic and interactive use is essential to fully understand the role of memes in online communication.

We lay the foundation by formalizing the Meme Reply Selection task, making the nuanced act of humorous replies computationally tractable by defining the necessary data properties and a systematic evaluation framework.
To support this task, we introduce the \textsc{MaMe-Re} benchmark illustrated in Figure~\ref{fig:overview}(c).  The quality and scale of the benchmark are ensured through a rigorous crowdsourcing process, yielding annotations from five workers for each of the 100{,}000 pairs. Also, its exhaustive pairing design, where every context is matched with every meme, allows for a unique analysis of how a single image's humorous potential changes with context. Finally, its openly licensed foundation ensures free redistribution and modification, fostering future research.

To demonstrate the challenges, our experiments show how current models handle contextual humor.
Our analysis reveals that methods using large language models (LLMs) show an emerging ability to capture complex social cues in humor.
Our qualitative analysis suggests that LLMs perform better because they can capture humorous strategies such as irony and exaggeration.
In contrast, embedding-based methods focus on surface-level semantic similarity, producing replies that are contextually relevant but often lack the element of surprise essential to humor.
Our analysis also identified several fundamental challenges.
Current models found it difficult to effectively leverage visual information, with performance sometimes degrading when it was provided.
Furthermore, while models perform well and can sometimes match human judgment when the humorous option is clearly distinguishable, they struggle with the more difficult and realistic task of selecting the best reply from a set of semantically similar candidates.
These results indicate that the core challenges for this task lie in understanding multimodal humor cues, as well as discerning the subtle differences in wit that make a reply truly effective.

The main contributions of this paper are as follows:
\begin{itemize}
    \item We propose treating memes as a form of interactive web communication, broadening the scope from the analysis of static artifacts to the dynamics of online conversation.
    \item We provide the foundational infrastructure for this paradigm: the formal Meme Reply Selection task and \textsc{MaMe-Re}, a benchmark featuring 100,000 human-annotated context-meme pairs, built entirely with openly licensed materials to ensure broad reusability.
    \item We present the first comprehensive analysis of this benchmark, identifying fundamental challenges for current models in understanding nuanced, multimodal humor.
\end{itemize}

\section{Related Work}
\label{sec:related_work}
Our research lies at the intersection of web communication, multimodal analysis, and computational humor. This section moves from the broad context of online social interaction down to the specific, unaddressed challenge of modeling humorous meme replies.

\paragraph{Web Communication and Multimodal Signals.}
Early studies of online interaction explored how users creatively adapt text-based media to convey rich socioemotional information, compensating for the lack of non-verbal cues \cite{Wal_1992, Wal_1996, Dresner2010FunctionsOT}. With the rise of social media, this adaptation has accelerated through the widespread integration of non-textual elements into digital conversations \cite{Highfield02012016}. Simple visual cues like emojis and GIFs have become integral to online expression, helping users manage interpersonal dynamics and convey subtle emotional nuances \cite{tianran-2017-spice-up, jiang-2018-perfect-one}. Memes represent a more advanced form of this multimodal practice and have evolved into a sophisticated medium for communication. They are used for a wide range of purposes, including not only humor~\cite{wu-2025-one-does-not, Zhong2023LetsTO} and hate speech~\cite{kiela-2020-hateful}, but also political statements~\cite{kulkarni2017internet, david-2020-the-evolution}, social commentary~\cite{alam-etal-2024-armeme}, and marketing~\cite{malodia-2022-meme-marketing}.

\paragraph{From Static Content to Interactive Replies.}
Computational research has predominantly treated memes as static artifacts for content analysis~\cite{sharma-etal-2020-semeval, pramanick-etal-2021-detecting, hossain-etal-2022-mute, wang-2024-memecraft, hwang-shwartz-2023-memecap}. This paradigm has proven valuable, advancing detection of harmful content~\cite{zhuang-2025-i-know}, understanding meme contexts~\cite{sharma-etal-2023-memex}, and generating humorous captions~\cite{zhang2024humor}. However, the focus on intrinsic properties of memes overlooks another key aspect of how memes function in online communication: as an interactive reply. In this view, the meaning and humorous effect are not contained within the meme itself, but are generated by the interaction between the meme and the conversational context.

\paragraph{Humor Mechanisms in Memes and Dialogue.}
The landscape of memes is diverse, ranging from reaction GIFs that convey a short story~\cite{wang-jurgens-2021-animated-picture} to photographic templates that derive new meaning from user-generated captions \cite{chen-2024-xmecap}. A useful lens for understanding meme replies is recontextualization~\cite{bauman1990poetics}, the act of taking a piece of media with a fixed meaning and placing it in a new conversational context to create a surprising juxtaposition. While some dialogue system research has explored selecting visual replies like stickers or emojis, these studies have primarily focused on semantic relevance~\cite{lu-2023-towards-building, wang-2019-culturally, wang-jurgens-2021-animated-picture}, rather than evaluating the humorous quality of the reply. This leaves the specific challenge of generating humorous replies largely unaddressed.

\section{Problem Formulation}
\label{sec:problem-formulation}
In this section, we formally define the task of meme reply selection,  specify the dataset requirements, and propose evaluation metrics for assessing model performance.

\paragraph{Task Definition.}
Let the context be a natural-language utterance $c \in \mathcal{C}$, where $\mathcal{C}$ is the set of arbitrary contexts in natural language.
Given a set of memes $\mathcal{M} = \{m_1, \dots, m_{|\mathcal{M}|}\}$, the task is to select the funniest meme for a given context:
\[
  \hat{m}(c) = \arg\max_{m \in \mathcal{M}} s(c, m),
\]
where $s(c, m)$ is a scoring function that measures the funniness of meme $m$ given the context $c$.

\paragraph{Dataset Requirements.}
To evaluate models for this task, we require a dataset $\mathcal{D}$ that consists of a finite set of contexts $\mathcal{C}_{\mathcal{D}} = \{c_1, \dots, c_{|\mathcal{C}_{\mathcal{D}}|}\} \subset \mathcal{C}$, a finite set of memes $\mathcal{M}_{\mathcal{D}} = \{m_1, \dots, m_{|\mathcal{M}_{\mathcal{D}}|}\} \subset \mathcal{M}$, and a real-valued scoring function $r : \mathcal{C}_{\mathcal{D}} \times \mathcal{M}_{\mathcal{D}} \rightarrow \mathbb{R}$, which assigns a funniness score to each context–meme pair.
Thus, the dataset is represented as a set of annotated triples:
\[
  \mathcal{D} = \{ (c, m, r(c, m)) \mid c \in \mathcal{C}_{\mathcal{D}},\; m \in \mathcal{M}_{\mathcal{D}} \}.
\]

\paragraph{Evaluation Metric.}
To evaluate model performance, we define Score@1 as the average reference funniness score of selected memes:
\[
  \mathrm{Score\text{@}1} =
  \frac{1}{|\mathcal{C}_{\mathcal{D}}|}
  \sum_{c \in \mathcal{C}_{\mathcal{D}}} r(c, \hat{m}(c)),
\]
where $r(c, m)$ denotes the reference funniness score from the dataset.
This metric directly evaluates our objective of selecting the most humorous meme reply for each given context.

\section{\textsc{MaMe-Re}: Manga Meme Reply Benchmark}
\label{sec:mamere-benchmark}
This section introduces \textsc{MaMe-Re}, a dataset for the meme reply selection task. It is designed to serve as a fundamental resource for this new research area. To ensure data consistency and reusability, we focused on a single meme domain, Japanese manga panels taken from a single copyright-free source, \textit{Black Jack ni Yoroshiku}.\footnote{\url{https://densho810.com/free/}. All data and annotations are in Japanese; English translations are provided for clarity.} The benchmark comprises 250 synthetically generated social media contexts and 400 manga panels, which form 100,000 unique context-meme pairs. Each pair was evaluated by five independent crowdworkers on a binary funniness scale, yielding a total of 500,000 annotations.

\paragraph{Content Collection and Curation.}
We created the benchmark content with a focus on reusability and variety within a specific domain.
We extracted 400 panels from \textit{Black Jack ni Yoroshiku} (\textit{Give My Regards to Black Jack} in English), provided by Sato Manga Works Ltd. under a free-use license that allows both commercial and non-commercial reproduction, public transmission, and a broad range of derivative uses, provided that attribution of the original title is maintained.
The panels cover a wide range of characters, scenes, and emotions.
For each panel, the in-panel speech text was transcribed, and GPT-4.1 was used to generate a visual description.
We also prepared 250 synthetic social media contexts, generated by GPT-4.1 to mimic posts on X (formerly Twitter) and screened for relevance and diversity.
We chose synthetic contexts instead of real posts to ensure privacy safety and open licensing, avoiding the ethical risks of scraping real tweets while maintaining the dataset's reusability.
We chose manga because its humor often arises from recontextualizing original scenes without alteration, enabling a focused study of humor as an emergent property of the interaction between meme and context.
Since both the images and the generated texts are based on freely licensed or synthetic content, the dataset is openly available and easy to reuse for further research.

\paragraph{Funniness Annotation.}

\begin{figure}[t]
  \centering
  \includegraphics[width=1.0\linewidth]{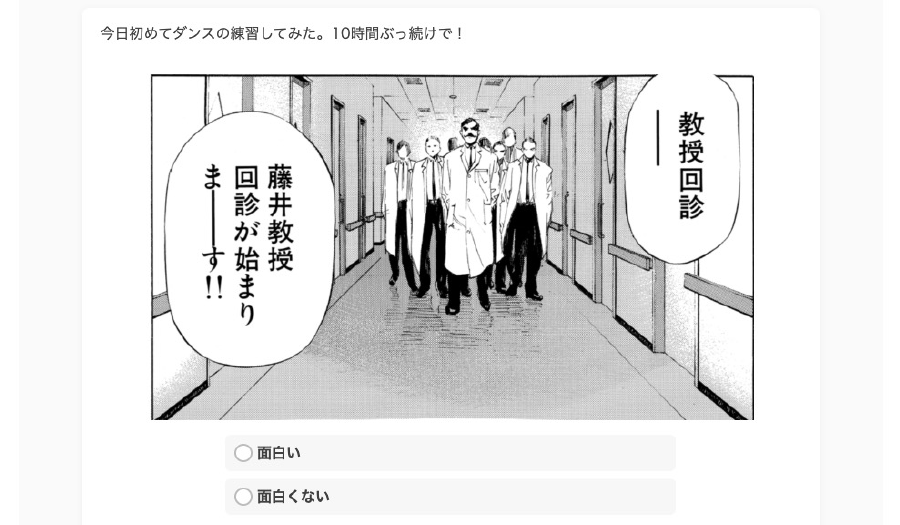}
  
  \tiny
  \begin{tcolorbox}[
      colback=white,
      colframe=black,
      boxrule=0.5pt,
      arc=0pt,
      left=2pt,
      right=2pt,
      top=2pt,
      bottom=2pt
  ]
\begin{Verbatim}
[Overview]
- This task involves evaluating the humor of image replies 
  to SNS posts.
- You will be shown an SNS post and an image reply to that post.
- Judge whether the image reply is funny or not as a response 
  to the SNS post.
- Please answer intuitively, without overthinking.
- Please answer as instructed for quality control questions.
- If your answers differ from the instructions on these 
  questions, your work may not be approved, 

(Example)

[Agreement]
${CONSENT_STATEMENT}
\end{Verbatim}
  \end{tcolorbox}
  \caption{\small Crowdworker annotation interface and full task instruction for the funniness scoring task in \textsc{MaMe-Re}. Top: interface screenshot. Bottom: instruction text shown to annotators.}
  \label{fig:annotation-interface-and-instructions}
\end{figure}
To obtain ground-truth labels for humor, we conducted a crowdsourcing task involving 2,325 unique annotators on Yahoo! Crowdsourcing\footnote{\url{https://crowdsourcing.yahoo.co.jp/}} in which annotators were shown a context–meme pair and asked to make a binary judgment: \textit{funny} or \textit{not funny}. The annotation interface displayed the context, meme image, and a binary choice, with full task instructions shown to annotators (Figure~\ref{fig:annotation-interface-and-instructions}). This controlled annotation setup allows us to isolate the humor signal from social factors such as post popularity or user influence. To ensure high-quality data, we included quality-control questions with explicit instructions and discarded submissions from annotators who failed them. Annotators were informed that responses would be used for academic research and provided consent through the platform interface. No personal or demographic information (e.g., age, gender, cultural background) was collected, as such information was not available through the crowdsourcing platform; only aggregate, anonymized responses were used.\footnote{Annotators were compensated at approximately 200 JPY per hour (10 JPY per batch of 50 annotations, with an average completion time of three minutes per batch). This rate was set with reference to the standard for simple questionnaire crowdsourcing tasks in Japan (typically 50-300 JPY per hour) and was two-four times higher than the compensation for other comparable tasks on the same platform. This study was deemed exempt from IRB review as minimal-risk research using only anonymized, aggregated responses with no collection of personal information.} The final funniness score for each pair is defined as the proportion of annotators who judged it to be funny. We adopted the binary format, simply asking whether a reply is funny or not, based on preliminary experiments.\footnote{We initially considered a multi-level Likert scale for graded humor ratings. However, pilot annotations revealed that distinguishing fine-grained degrees of funniness was difficult for annotators, resulting in inconsistent and noisy labels. The higher cognitive load required for ranking also appeared to degrade annotation quality. In contrast, the binary judgment of funny or unfunny is intuitive and easy to answer, and aligns well with our goal of selecting a funny reply. Given these findings and the exploratory nature of this benchmark, we opted for the simpler binary approach.}

\paragraph{Dataset Statistics.}
The \textsc{MaMe-Re} dataset consists of 250 contexts and 400 manga panels, forming 100,000 unique context-meme pairs with five annotations each, yielding a total of 500,000 annotations. The overall inter-annotator agreement, measured by Fleiss' $\kappa$, is -0.022, reflecting the profound subjectivity of humor. However, this low overall agreement does not imply that the annotations are random noise. Crucially, a substantial portion of the data reached high levels of consensus: 20.4\% of pairs achieved unanimous agreement (a score of 0.0 or 1.0), and an additional 21.8\% had high agreement (a score of 0.2 or 0.8). This demonstrates that while many pairs are subjective, a significant subset elicits consistent judgments, providing a solid foundation for evaluating models.

\section{Reply Selection Methods}
\label{sec:reply-selection-methods}
We investigate two different strategies for meme reply selection: (i) similarity-based selection, which relies on indirect semantic matching, and (ii) preference-based selection, which directly models funniness given the context.

\paragraph{Similarity-based Selection (\texttt{sim-select}).}
This approach selects memes by measuring cosine similarity between the context and each candidate meme in embedding space. 
This score serves as an indirect proxy for funniness, relying on the assumption that similar memes are more likely to be suitable and funny replies.
Specifically, we use a text encoder $f_c$ to obtain $f_c(c) \in \mathbb{R}^d$ for the context, and text or multimodal encoders $f_m$ to obtain $f_m(m) \in \mathbb{R}^d$ for the memes. The meme with the highest cosine similarity to the context is selected:

\[
\hat{m}(c) = \arg\max_{m \in \mathcal{M}} \cos\bigl(f_c(c), f_m(m)\bigr).
\]

The embedding $f_m(m)$ can use speech text, panel description, or the image.
In our experiments, we implement $f_m$ with four model types: text embeddings (Text-Emb), LLM embeddings (LLM-Emb), multimodal embeddings (Multi-Emb), and vision-language model embeddings (VLM-Emb). For text inputs, we use either speech text alone or with the panel description. Multi-Emb uses only the image, while VLM-Emb incorporates both speech text and panel image.

\paragraph{Preference-based Selection (\texttt{pref-select}).}
Preference-based selection directly estimates the funniness of each candidate given the context, rather than relying on vector proximity.
Specifically, given a context $c$ and meme set $\mathcal{M}$, a preference function $f_p$ evaluates candidates and selects the meme judged funniest by the model: 

\[
\hat{m}(c) = f_p(c, \mathcal{M}).
\]

In our experiments, we implement $f_p$ using LLMs with prompts instructing the model to choose the funniest meme (Figure~\ref{fig:prompt-template}), corresponding to the standard multiple-choice question answering~\cite{hendrycks2021measuring, robinson2023leveraging}. For meme information in the prompts, we use either speech text alone or with the panel description, mirroring the \texttt{sim-select} setting.

\begin{figure}[t]
  \tiny
  \begin{tcolorbox}[
      colback=white,
      colframe=black,
      boxrule=0.5pt,
      arc=0pt,
      left=2pt,
      right=2pt,
      top=2pt,
      bottom=2pt
  ]
  \begin{Verbatim}
  [Instructions]
  You will be given a post and a set of meme candidates.
  Select the funniest reply to the post from the candidates.
  Meme candidates appear in CSV with the columns: ${FORMAT}
  Choose the single option that will be the funniest.
  
  [Output Format]
  Output only the ID.
  Do not include any extra wording or markup.
  
  [Candidates]
  ${CANDIDATES}
  
  --------------
  post: {context}
  \end{Verbatim}
  \end{tcolorbox}
  \caption{\small Prompt template. \$\{FORMAT\} has ``\textit{id, speech}'' or ``\textit{id, speech, description}'' and \$\{CANDIDATES\} have meme candidates in the corresponding csv format.}
  \label{fig:prompt-template}
  \end{figure}

\section{Experiments}
\label{sec:experiments}
Our experiments are designed as a sequential investigation to identify the core challenges in humorous meme reply selection. We begin in Exp1 with a broad comparison of \texttt{sim-select} and \texttt{pref-select} methods to establish baseline performance. Finding that \texttt{pref-select} is promising but limited, we test a standard retrieve-and-rerank architecture in Exp2, which surprisingly fails to improve performance. To understand this failure, Exp3 simplifies the task to test if models can select a clearly humorous meme from a set of non-humorous ones. After confirming that the models can, Exp4 directly tests our final hypothesis: that performance drops when models must choose from a pool of semantically similar candidates. Finally, our case study provides a qualitative analysis of the models' humor generation strategies.

\subsection{Overall Settings}
\label{sec:overall-settings}
We used Score@1 as our primary evaluation metric, which directly measures how often the system selects a meme reply judged to be funny in each context. To account for the inherent subjectivity of humor, we also report the Consensus Hit Rate (CHR), defined as the rate of selecting memes that all five annotators unanimously judged as funny, and nDCG to evaluate model performance from a retrieval and ranking perspective. For similarity-based selection, we evaluated 21 model variants across four model types, focusing on Japanese-optimized models. In this section, we report results from the top-performing models in each category: Sarashina-Text-Emb and OpenAI-Text-Emb for Text-Emb; Calm2-LLM-Emb and Qwen2.5-LLM-Emb for LLM-Emb; LAION-CLIP-Multi-Emb and Jina-CLIP-Multi-Emb for Multi-Emb; EvoVLM-VLM-Emb and Sarashina-VLM-Emb for VLM-Emb. For preference-based selection, we used six production-grade LLMs: GPT-5, GPT-5-mini, GPT-OSS, Gemini 2.5 Pro, Gemini 2.5 Flash, and Claude 4 Sonnet.\footnote{Model details: sarashina-embedding-v1-1b and text-embedding-3-large for Text-Embs, CLIP-ViT-H-14-frozen-xlm-roberta-large-laion5B-s13B-b90k and jina-clip-v2 for Multimodal-Embs, Calm2-7b and Qwen2.5-7B for LLM-Embs, and sarashina2-vision-8b and Llama-3-EvoVLM-JP-v2 for VLM-Embs. gpt-5-2025-08-07, gpt-5-mini-2025-08-07, gpt-oss-120b, gemini-2.5-flash, gemini-2.5-pro, and claude-sonnet-4-20250514 for LLMs. For compute resources, preference-based selection used commercial APIs (OpenAI, Google, Anthropic), while similarity-based models were run on a single NVIDIA A100 GPU.} 

\subsection{Exp1: Main Results}
\label{sec:exp1}
\begin{figure*}[t]
  \centering
  
  \begin{minipage}{0.68\linewidth}
    \begin{subfigure}[b]{\linewidth}
      \centering
      \small
      \begin{tabular}{lcccl}
      \toprule
      Model & Desc & Score@1 & CHR & nDCG@5 \\
      \midrule
      \textbf{GPT-5} (P) & \textbf{Y} & \textbf{0.325 ($\pm$0.014)} & \textbf{0.052 ($\pm$0.016)} & - \\
      GPT-5 mini (P) & N & 0.322 ($\pm$0.014) & 0.045 ($\pm$0.015) & - \\
      \textbf{Sarashina-Text-Emb} (S) & \textbf{N} & \textbf{0.320 ($\pm$0.022)} & 0.024 ($\pm$0.019) & 0.317 \\
      GPT-5 (P) & N & 0.314 ($\pm$0.014) & 0.048 ($\pm$0.015) & - \\
      OpenAI-Text-Emb (S) & N & 0.307 ($\pm$0.024) & 0.032 ($\pm$0.022) & 0.321 \\
      \textbf{Calm2-LLM-Emb} (S) & \textbf{Y} & \textbf{0.306 ($\pm$0.024)} & 0.028 ($\pm$0.020) & 0.284 \\
      GPT-5 mini (P) & Y & 0.300 ($\pm$0.013) & 0.024 ($\pm$0.011) & - \\
      Qwen2.5-LLM-Emb (S) & Y & 0.299 ($\pm$0.023) & 0.032 ($\pm$0.022) & 0.278 \\
      Gemini 2.5 Flash (P) & N & 0.296 ($\pm$0.014) & 0.029 ($\pm$0.012) & - \\
      Calm2-LLM-Emb (S) & N & 0.295 ($\pm$0.023) & 0.024 ($\pm$0.019) & 0.294 \\
      Gemini 2.5 Pro (P) & Y & 0.294 ($\pm$0.014) & 0.033 ($\pm$0.013) & - \\
      GPT-OSS (P) & N & 0.292 ($\pm$0.014) & 0.035 ($\pm$0.013) & - \\
      Gemini 2.5 Pro (P) & N & 0.286 ($\pm$0.014) & 0.032 ($\pm$0.013) & - \\
      OpenAI-Text-Emb (S) & Y & 0.285 ($\pm$0.024) & 0.020 ($\pm$0.017) & 0.314 \\
      Sarashina-Text-Emb (S) & Y & 0.285 ($\pm$0.023) & 0.012 ($\pm$0.013) & 0.294 \\
      GPT-OSS (P) & Y & 0.284 ($\pm$0.014) & 0.024 ($\pm$0.011) & - \\
      \textbf{Jina-CLIP-Multi-Emb} (S) & \textbf{N} & \textbf{0.282 ($\pm$0.025)} & 0.020 ($\pm$0.017) & 0.294 \\
      LAION-CLIP-Multi-Emb (S) & N & 0.280 ($\pm$0.025) & 0.028 ($\pm$0.020) & 0.289 \\
      Claude 4 (P) & N & 0.280 ($\pm$0.013) & 0.041 ($\pm$0.046) & - \\
      Qwen2.5-LLM-Emb (S) & N & 0.279 ($\pm$0.023) & 0.016 ($\pm$0.016) & 0.289 \\
      Gemini 2.5 Flash (P) & Y & 0.276 ($\pm$0.014) & 0.023 ($\pm$0.011) & - \\
      \textbf{EvoVLM-VLM-Emb} (S) & \textbf{N} & \textbf{0.263 ($\pm$0.022)} & 0.012 ($\pm$0.013) & 0.262 \\
      Claude 4 (P) & Y & 0.261 ($\pm$0.013) & 0.000 ($\pm$0.000) & - \\
      Sarashina-VLM-Emb (S) & N & 0.261 ($\pm$0.023) & 0.020 ($\pm$0.017) & 0.280 \\ 
      \midrule
      Random & - & 0.255 ($\pm$0.001) & 0.015 ($\pm$0.015) & 0.271 \\
      \bottomrule
      \end{tabular}
      \caption{Performance ranking (Exp1).}
      \label{tab:sub-results-side}
    \end{subfigure}
  \end{minipage}
  \hfill
  \begin{minipage}{0.3\linewidth}
    \centering
    \begin{subfigure}[b]{\linewidth}
      \centering
      \includegraphics[width=\linewidth]{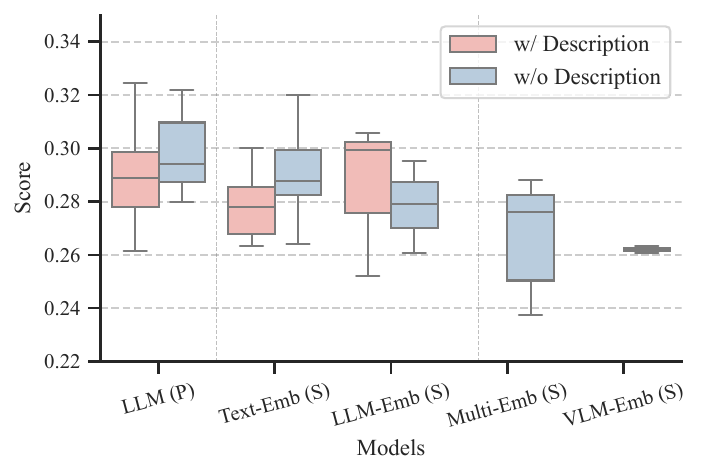}
      \caption{By description}
      \label{fig:sub-desc-side}
    \end{subfigure}
    \vspace{0.3em}
    \begin{subfigure}[b]{\linewidth}
      \centering
      \includegraphics[width=\linewidth]{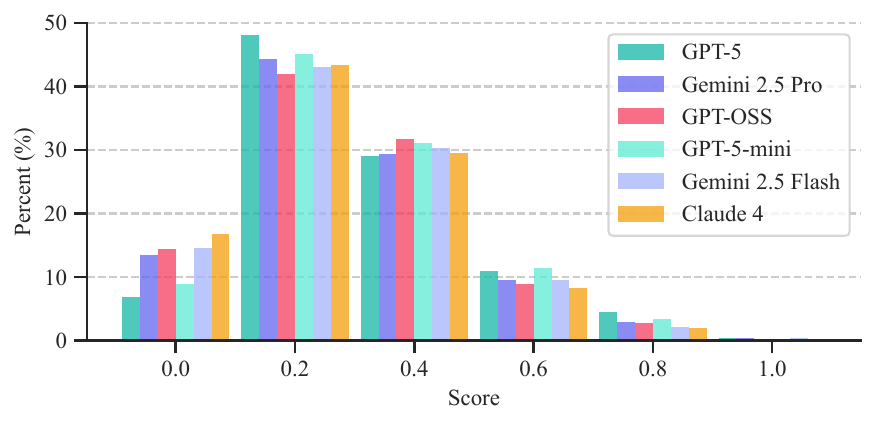}
      \caption{By LLMs}
      \label{fig:sub-llm-side}
    \end{subfigure}
    \vspace{0.3em}
    \begin{subfigure}[b]{\linewidth}
      \centering
      \includegraphics[width=\linewidth]{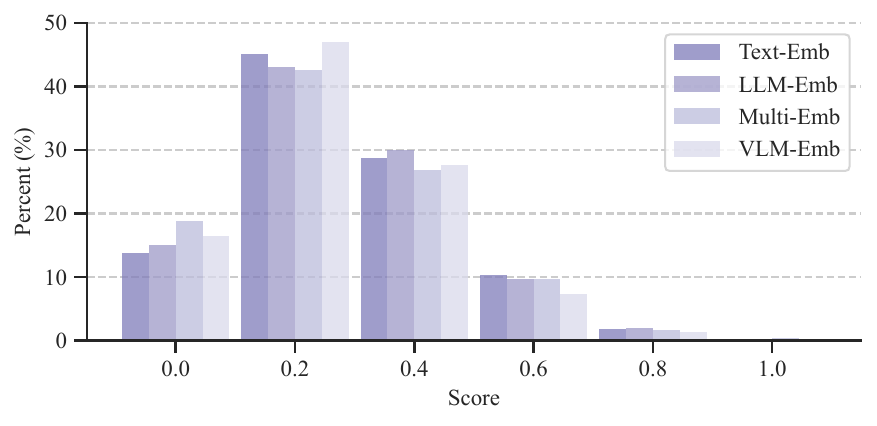}
      \caption{By embeddings}
      \label{fig:sub-emb-side}
    \end{subfigure}
  \end{minipage}

  \vspace{0.5em}
  \caption{Main experimental results for Exp1. (a) \textbf{Table} showing the performance ranking across models and methods. S/P: similarity/preference-based; Y/N: with/without descriptions; CHR: Consensus Hit Rate; values in parentheses denote 95\% confidence intervals. (b)--(d) \textbf{Plots} of score distributions categorized by panel descriptions, LLMs, and embedding models respectively.}
  \label{fig:combined-side-layout}
\end{figure*}

Figure~\ref{tab:sub-results-side} shows the results of Exp1. The \texttt{pref-select} method using GPT-5 achieved a Score@1 of 0.325 when using panel descriptions. For \texttt{sim-select} methods, Sarashina-Text-Emb achieved 0.320 without descriptions. Calm2-LLM-Emb (LLM-Emb) achieved 0.306, while multimodal and vision-language models showed lower scores: Jina-CLIP-Multi-Emb (0.282) and EvoVLM-VLM-Emb (0.263). This pattern reveals that methods capable of reasoning about contextual humor (\texttt{pref-select}) outperform those relying on semantic similarity alone, and that visual information does not consistently improve performance.
To assess the ranking quality of the \texttt{sim-select} methods, we also calculated nDCG@5. While these scores were broadly consistent with the Score@1 rankings, the performance differences among the top models were marginal.


Figure~\ref{fig:sub-desc-side} shows the effect of including textual panel descriptions on model performance. Overall, the results indicate that providing these descriptions did not consistently improve scores. For the \texttt{pref-select} and Text-Emb models, performance slightly decreased when descriptions were included. Only the LLM-Emb models showed a minor benefit. Since the score ranges with and without descriptions largely overlap, this additional text offered limited benefit. This finding, combined with the poor performance of models that directly process images, highlights a significant challenge: current models struggle to use visual information effectively, whether it is provided as text or as pixels.

To address concerns about annotation reliability raised by low overall agreement, we evaluated models on a high-agreement subset where ground truth is defined strictly as pairs with a funniness score $\geq 0.8$ (unanimous or near-unanimous agreement). The Consensus Hit Rate (CHR) (the proportion of universally funny memes selected by each model) is shown in Figure~\ref{tab:sub-results-side}. GPT-5 with description achieved the highest consensus hit rate of 0.052 ($\pm$0.016), which is approximately 3.5 times higher than the random baseline (0.015) and 2.2 times higher than Sarashina-Text-Emb without description (0.024, $\pm$0.019). This demonstrates that LLMs maintain their advantage even when evaluated on the most reliable subset of annotations, validating our findings in a noise-free setting. However, the absolute performance remains low (only 5.2\% of universally funny memes are selected), highlighting the persistent challenge of humor detection.



To better understand model behavior, we analyzed the distribution of funniness scores for the memes selected by each model.
Figure~\ref{fig:sub-llm-side} shows the score distributions for the LLMs used in the \texttt{pref-select} approach. GPT-5 and GPT-5-mini effectively avoided completely unfunny memes (score of 0.0), selecting them less than 10\% of the time. These models also selected funny memes (scores > 0.6) more often than the other LLMs. In contrast, Claude 4 was more likely to choose memes with a score of zero and less likely to select highly rated ones. These different patterns suggest that LLMs have varying abilities to understand and select for humor.

Figure~\ref{fig:sub-emb-side} presents the distributions for the \texttt{sim-select} models. The Text-Emb and LLM-Emb models show an advantage in avoiding zero-score memes compared to multimodal models. At the higher end of the funniness scale, the differences were less pronounced, though Text-Emb and LLM-Emb models performed slightly better. This suggests that text-based similarity is effective at capturing enough semantic relevance to avoid completely inappropriate replies. Conversely, the frequent selection of such memes by multimodal models may indicate a fundamental difficulty in grasping contextual relevance from visual information for this task.

\subsection{Exp2: Retrieve-and-Rerank Approach}
\label{sec:exp2}

\paragraph{Settings}
In this experiment, we examined a hybrid retrieve-and-rerank approach, combining the \texttt{sim-select} and \texttt{pref-select} methods. This method first utilizes \texttt{sim-select} to retrieve the top-k similar candidates for a given context. Subsequently, a \texttt{pref-select} model is employed to identify the most humorous option from this reduced set. For the retrieval step, we used Sarashina-Text-Emb, the top-performing similarity-based model from Exp1. For the selection step, we evaluated three large language models chosen from our Exp1 results to represent a range of top-performing proprietary and open models: GPT-5, Gemini 2.5 Pro, and GPT-OSS. To better interpret their performance, we established two baselines: Random, which randomly selects a meme from the $k$ candidates, and Oracle, which serves as an oracle upper bound by always choosing the candidate with the highest ground-truth funniness score within the retrieved set.

\paragraph{Results}
\begin{figure}
    \centering
    \includegraphics[width=1.0\linewidth]{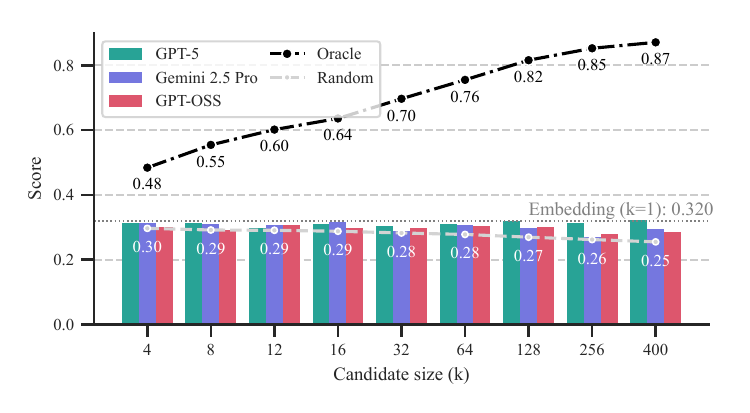}
    \caption{Performance of the retrieve-and-rerank approach (Exp2).}
    \label{fig:rag_results}
\end{figure}

Figure~\ref{fig:rag_results} shows the results of our retrieve-and-rerank approach. A key observation is the notable gap between the oracle upper bound and the actual performance of the \texttt{pref-select} models. The rising oracle baseline indicates that the initial retrieval phase successfully included high-quality candidates in the selection pool, even for small values of $k$; for example, Score@1 reached 0.48 at $k=4$ and 0.55 at $k=8$. However, the models did not effectively leverage this advantage.

The performance of the \texttt{pref-select} models remained near or below the dashed line representing the standalone \texttt{sim-select}  baseline of 0.320. This suggests that adding a \texttt{pref-select} step did not yield a consistent improvement over the simpler retrieval method. The models showed different behaviors as the candidate pool size $k$ varied. The performance of GPT-5 was highly stable, with scores showing little variation around 0.31 regardless of $k$. In contrast, Gemini 2.5 Pro and GPT-OSS showed a tendency for their performance to degrade as $k$ increased. Specifically, their scores dropped at $k\geq256$.

This result indicates a mixed pattern of behavior. The stability of GPT-5 suggests that for some models, the task's difficulty is not a function of the number of candidates, which appears to contradict the common \emph{lost-in-the-middle} problem in LLMs~\cite{li-etal-2024-loogle, liu-etal-2024-lost}. Conversely, the performance decline of other models suggests a degree of sensitivity to context length. Overall, these results suggest that the limited improvement may not stem from the retrieval process itself, but rather from the difficulty of distinguishing subtle differences in humor among semantically similar candidates.

\subsection{Exp3: Distinguishable Candidate Selection}
\label{sec:exp3}
\paragraph{Settings}
To better understand why the retrieve-and-rerank approach from Exp2 was limited, we designed a controlled experiment with a simplified task. The goal was to test two basic hypotheses under controlled conditions: (1) that models can reliably select a distinctly humorous meme from a set of less funny alternatives, and (2) that performance would decrease as the number of candidates $k$ increases. To do this, we constructed test sets where each contained one funny meme (score $\ge$ 0.8), half the candidates ($k/2$) were unfunny memes ($\text{score} \le 0.2$), and the remaining ones were somewhat funny ($0.2 < \text{score} < 0.8$).

In addition to the models from our previous experiments, we included two additional baselines for a comprehensive comparison. First, we collected human performance data through a crowdsourced selection task to serve as a practical reference. Second, to assess the role of visual understanding, we evaluated vision-language models (GPT-5 Vision, Gemini 2.5 Pro Vision). For these models, all candidate images for a given context were combined into a single grid image, and the model was prompted to select the best option from this composite image.

\paragraph{Results}
\begin{figure}
    \centering
    \includegraphics[width=1.0\linewidth]{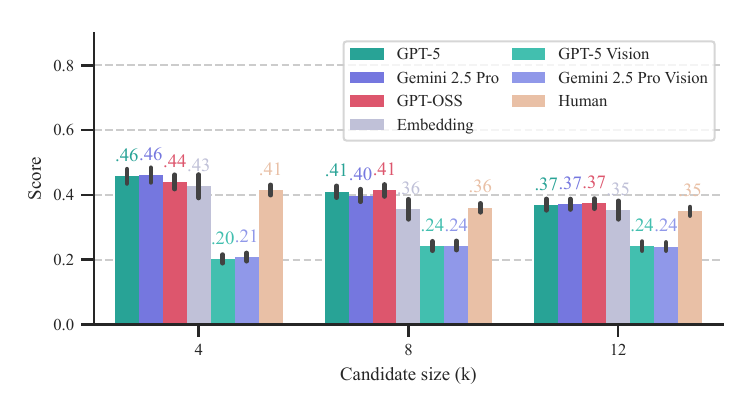}
    \caption{\small Performance on selecting from distinguishable candidates (Exp3). Human and vision-language models are included as baselines.}
    \label{fig:human_llm_performance}
\end{figure}

As shown in Figure~\ref{fig:human_llm_performance}, the results from this controlled experiment were consistent with the hypotheses. First, the LLMs were effective at identifying the target funny meme, with scores up to 0.46. Second, unlike in Exp2, there was a clear decrease in performance as $k$ increased for both the models and the human reference. For instance, the scores for GPT-5 were 0.46, 0.41, and 0.37 for $k$=4, 8, and 12, respectively.

While the artificial setting of this experiment limits its real-world applicability, it reveals important aspects of model behavior. The results show that when a humorous option was clearly distinguishable, the \texttt{pref-select} models were close to, and in some cases slightly higher than, the average human performance on this task. This success was not replicated by the vision-language models, whose poor performance indicated that a deep understanding of visual humor remains a challenge. The experiment highlights a key distinction: while models can approach human performance on tasks with clear signals, their ability to navigate more complex, real-world scenarios has not yet been demonstrated.

\subsection{Exp4: Sensitivity to Candidate Similarity}
\label{sec:exp4}

\paragraph{Settings}
In Exp3, we observed that LLMs performed well when funny memes were clearly distinguishable from less funny ones. Building on this observation, we examined how LLM performance changes when the candidate memes become more similar to each other, extending the controlled setting of Exp3. This follows the implication from Exp2 that the retrieve-and-rerank approach may have been limited by the high similarity among retrieved candidates, creating a \emph{hard negative} scenario that makes it difficult for the reranker to identify the single best option.

To test this hypothesis, we modified the evaluation sets from Exp3. We retained the funny candidates ($\text{score} \ge 0.8$) and resampled the other candidates based on their similarity to the funny one, using the similarity threshold $t$. We selected only candidates whose similarity score with the funny meme exceeded this threshold. This approach allowed us to control the similarity within the candidate pool. We used three thresholds ($t$=\{-2,0,2\}), based on the standardized cosine similarity from Sarashina-Text-Emb embeddings, denoting these conditions as Low, Mid, and High, respectively. The Low condition contains a diverse set of candidates, whereas the High condition requires the model to distinguish the funny meme from a pool of semantically similar yet unfunny alternatives.

\paragraph{Results}

\begin{table}[t]
  \centering
  \small
\begin{tabular}{llcccc} 
\toprule
    \multirow{2}{*}{$k$} & \multirow{2}{*}{Model} & \multicolumn{3}{c}{Candidate Similarity} & \multirow{2}{*}{$\Delta$} \\ \cmidrule{3-5}
     & &     Low &     Mid & High &   \\
\midrule
    \multirow{3}{*}{4} 
     & GPT-5          & 0.457 & 0.459 & 0.437 & -0.020 \\
     & Gemini 2.5 Pro & 0.467 & 0.468 & 0.444 & -0.023 \\
     & GPT-OSS        & 0.457 & 0.456 & 0.385 & -0.072 \\
\midrule
    \multirow{3}{*}{8} 
     & GPT-5          & 0.396 & 0.416 & 0.379 & -0.017 \\
     & Gemini 2.5 Pro & 0.389 & 0.400 & 0.375 & -0.014 \\
     & GPT-OSS        & 0.402 & 0.386 & 0.348 & -0.054 \\
\midrule
  \multirow{3}{*}{12} 
       & GPT-5          & 0.365 & 0.365 & 0.356 & -0.009 \\
   & Gemini 2.5 Pro & 0.370 & 0.371 & 0.356 & -0.014 \\
   & GPT-OSS        & 0.371 & 0.355 & 0.341 & -0.030 \\
\bottomrule
  \end{tabular}
  \caption{\small LLM performance under different candidate similarities. $\Delta$ represents the performance change between the Low and High similarity conditions. (Exp4).}
  \label{tab:similarity-threshold-results}
\end{table}

Table~\ref{tab:similarity-threshold-results} shows the results. As hypothesized, LLM performance generally decreased as the similarity between candidates increased. This effect was most pronounced for smaller candidate pools ($k$=4), suggesting that high similarity makes it more difficult for the models to identify the funniest meme. As $k$ increased, the performance drop became smaller. This suggests that the challenge of processing a larger number of candidates may outweigh the difficulty caused by their similarity. This effect was also stronger for GPT-OSS, which, as a less powerful model, was more affected by this hard negative scenario.

These findings suggest that LLMs struggle to distinguish subtle differences in humor among semantically similar candidates. This observation offers a plausible explanation for the limited performance of the retrieve-and-rerank approach in Exp2. It is possible that the initial retrieval step in that experiment created a pool of highly similar candidates, making it difficult for LLMs to discern the most humorous option.

\subsection{Case Study}
\label{sec:case-study}
\begin{figure}
    \centering
    \includegraphics[width=1.0\linewidth]{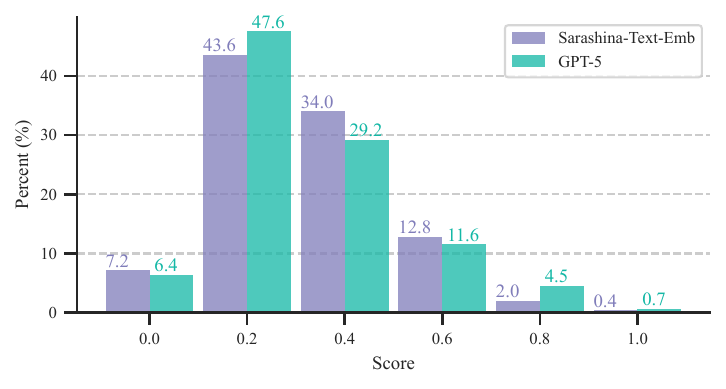}
    \caption{\small Score distributions of Sarashina-Text-Emb and GPT-5.}
    \label{fig:score-distributions-sarashina-gpt5}
\end{figure}

This section analyzes how the \texttt{sim-select} and \texttt{pref-select} approaches create humor by examining their patterns of success and failure. We focus on the outputs of Sarashina-Text-Emb and GPT-5, the best-performing models from each method. An analysis of their overall performance in Figure~\ref{fig:score-distributions-sarashina-gpt5} reveals two distinct behavioral patterns. The approach of Sarashina-Text-Emb is reliable but conservative; it tended to select memes that are moderately funny but semantically close to the context. In contrast, GPT-5 showed a high-risk, high-reward pattern, more frequently selecting very funny replies. To understand the mechanisms that produce these different outcomes, we now examine specific examples in Table~\ref{tab:examples}.

\begin{table}
  \centering
  \small
  \hrule height 0.6pt
  \vspace{1mm}

  \noindent 
  \begin{minipage}[b]{0.33\linewidth} 
    \centering
    Context
  \end{minipage}%
  \hfill 
  \begin{minipage}[t]{0.3\linewidth} 
  \centering
    Sarashina-Text-Emb \\ (Similarity-based)
  \end{minipage}%
  \hfill
  \begin{minipage}[t]{0.33\linewidth} 
  \centering
    GPT-5 \\ (Preference-based)
  \end{minipage}

  \vspace{1mm}
  \hrule height 0.4pt
  \vspace{2mm}

  \noindent 
  \begin{minipage}[t][2.5cm][c]{0.33\linewidth} 
    (A) \textit{\footnotesize The Wi-Fi at my new place is seriously fast!}  \\
  \end{minipage}%
  \hfill
  \begin{minipage}[t][2.5cm][t]{0.3\linewidth}
    \textcolor{green}{\Large\ding{51}} \small{(0.8)} \\
    \includegraphics[width=\linewidth, height=0.8\textheight, keepaspectratio]{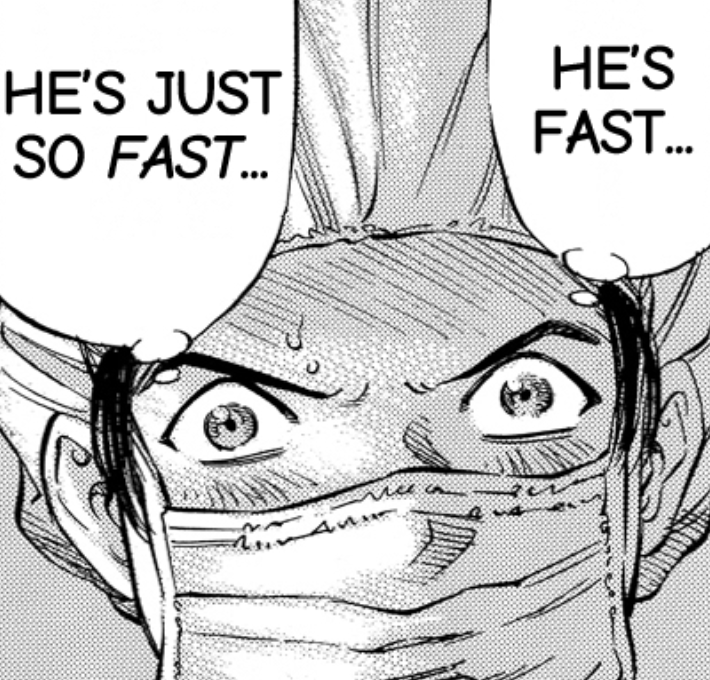}
  \end{minipage}%
  \hfill
  \begin{minipage}[t][2.5cm][t]{0.3\linewidth}
    \textcolor{green}{\Large\ding{51}} \small{(0.8)} \\
    \includegraphics[width=\linewidth, height=0.8\textheight, keepaspectratio]{figures/both_funny.png}
  \end{minipage}
  \vspace{2mm}
  \hrule height 0.4pt 
  \vspace{2mm}

  \noindent 
  \begin{minipage}[t][2cm][c]{0.33\linewidth}
    (B) \textit{\footnotesize I think I'm lost...}  \\
  \end{minipage}%
  \hfill
  \begin{minipage}[t][2cm][t]{0.3\linewidth}
    \textcolor{green}{\Large\ding{51}} \small{(1.0)} \\
    \includegraphics[width=\linewidth, height=0.8\textheight, keepaspectratio]{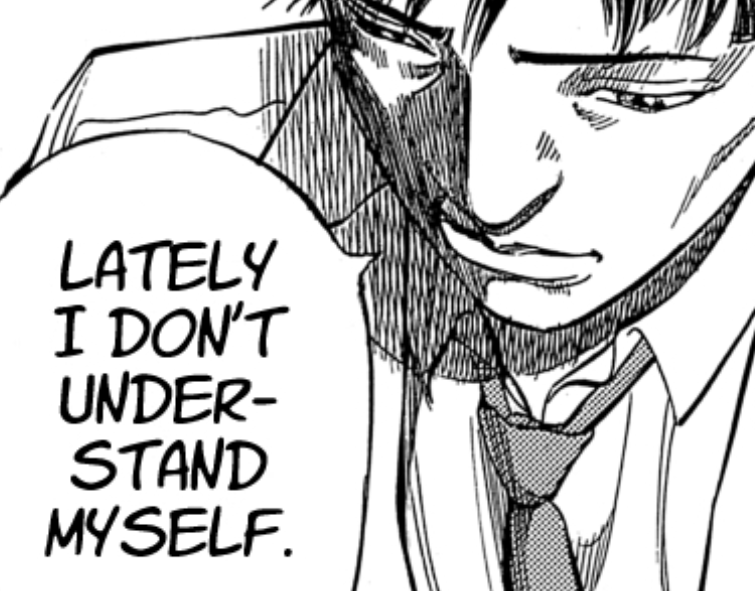}
  \end{minipage}%
  \hfill
  \begin{minipage}[t][2cm][t]{0.3\linewidth}
    \textcolor{red}{\Large\ding{55}} \small{(0.2)} \\
    \includegraphics[width=\linewidth, height=0.8\textheight, keepaspectratio]{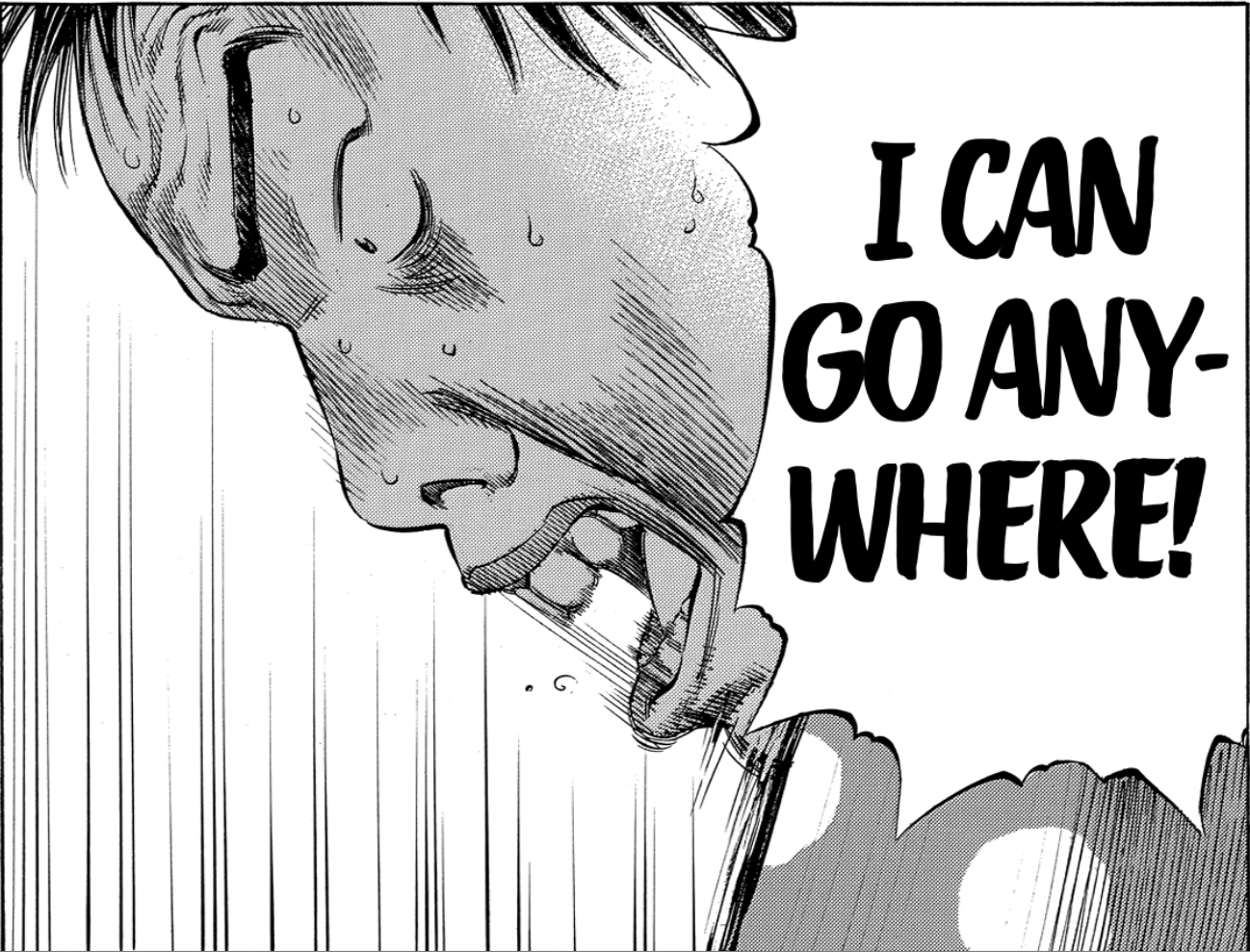}
  \end{minipage}

  \vspace{2mm}
  \hrule height 0.4pt
  \vspace{2mm}

  \noindent 
  \begin{minipage}[t][2cm][c]{0.33\linewidth}
    (C) \textit{\footnotesize I'm done. I give up.}  \\
  \end{minipage}%
  \hfill
  \begin{minipage}[t][2cm][t]{0.3\linewidth}
  \textcolor{green}{\Large\ding{51}} \small{(0.6)} \\
    \includegraphics[width=\linewidth, height=0.8\textheight, keepaspectratio]{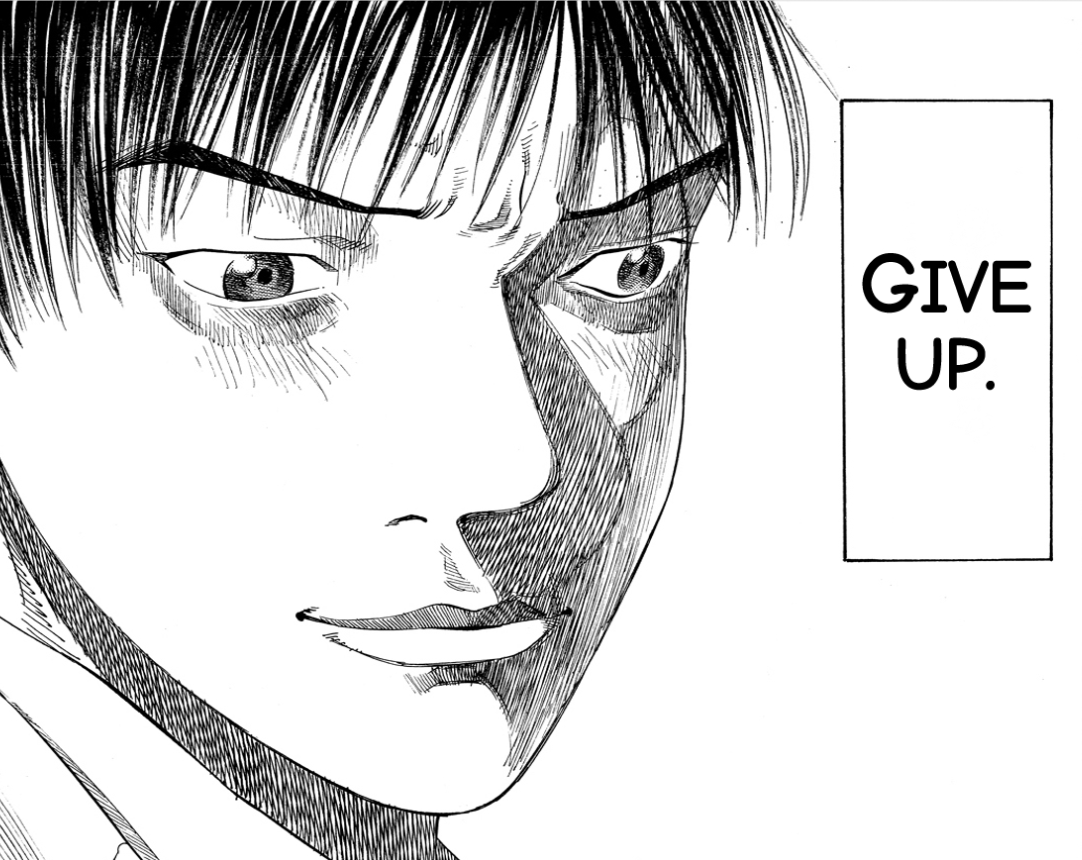}
  \end{minipage}%
  \hfill
  \begin{minipage}[t][2cm][t]{0.3\linewidth}
    \textcolor{red}{\Large\ding{55}} \small{(0.2)} \\
    \includegraphics[width=\linewidth, height=0.8\textheight, keepaspectratio]{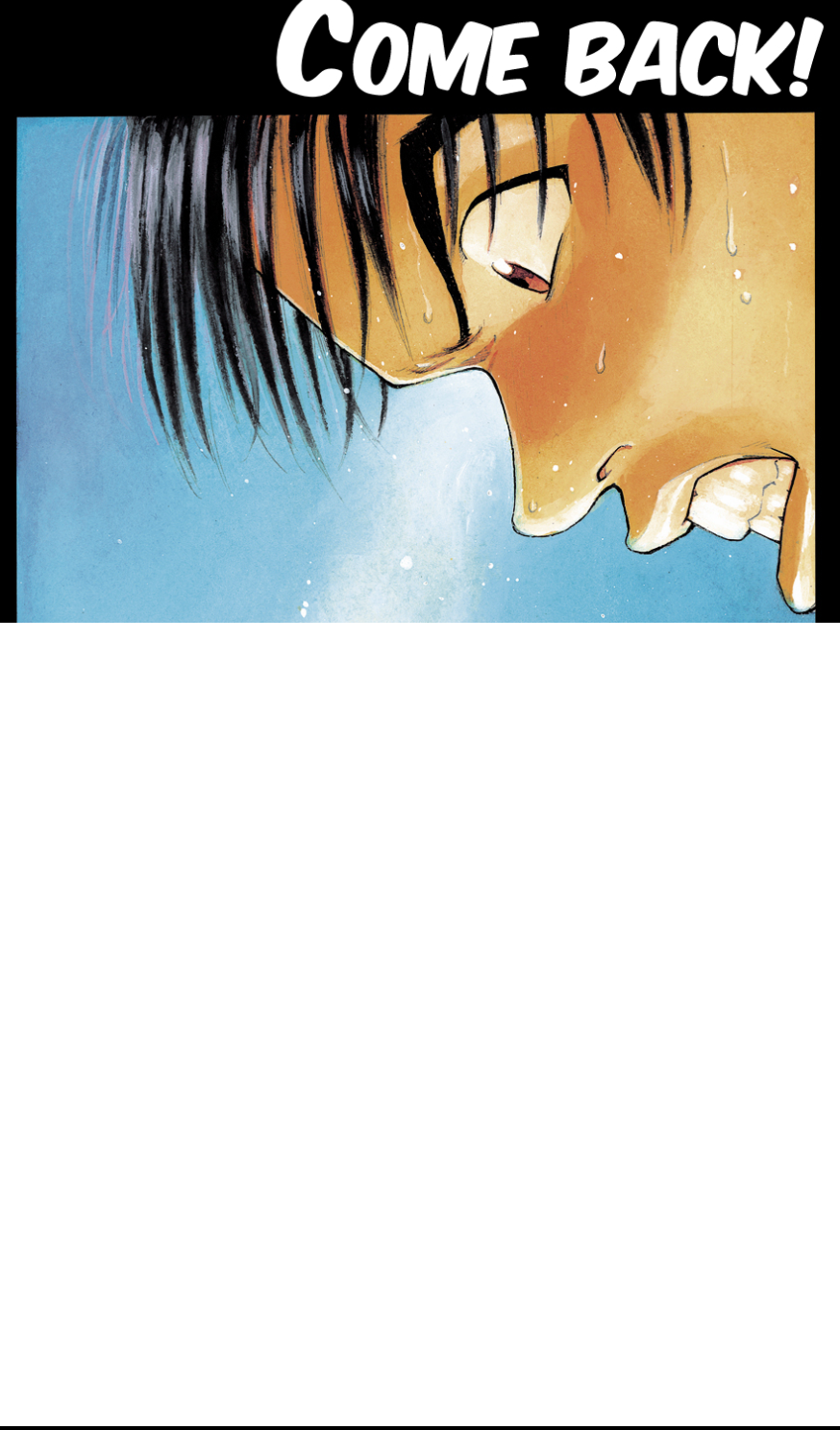}
  \end{minipage}
  
  \vspace{2mm}
  \hrule height 0.4pt
  \vspace{2mm}
  
  \noindent 
  \begin{minipage}[t][1.5cm][c]{0.33\linewidth}
    (D) \textit{\footnotesize Just found a 10-year-old pudding in the back of my fridge...}  \\
  \end{minipage}%
  \hfill
  \begin{minipage}[t][1.5cm][t]{0.3\linewidth}
  \textcolor{red}{\Large\ding{55}} \small{(0.2)} \\
    \includegraphics[width=\linewidth, height=0.8\textheight, keepaspectratio]{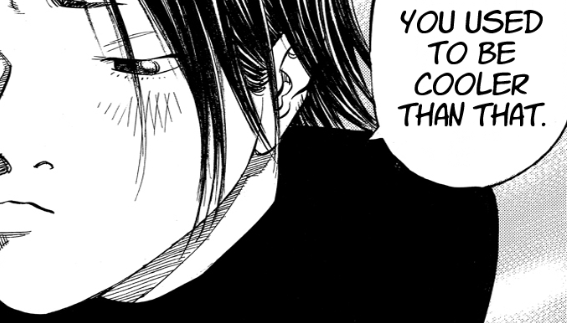}
  \end{minipage}%
  \hfill
  \begin{minipage}[t][1.5cm][t]{0.3\linewidth}
    \textcolor{green}{\Large\ding{51}} \small{(0.8)} \\
    \includegraphics[width=\linewidth, height=0.8\textheight, keepaspectratio]{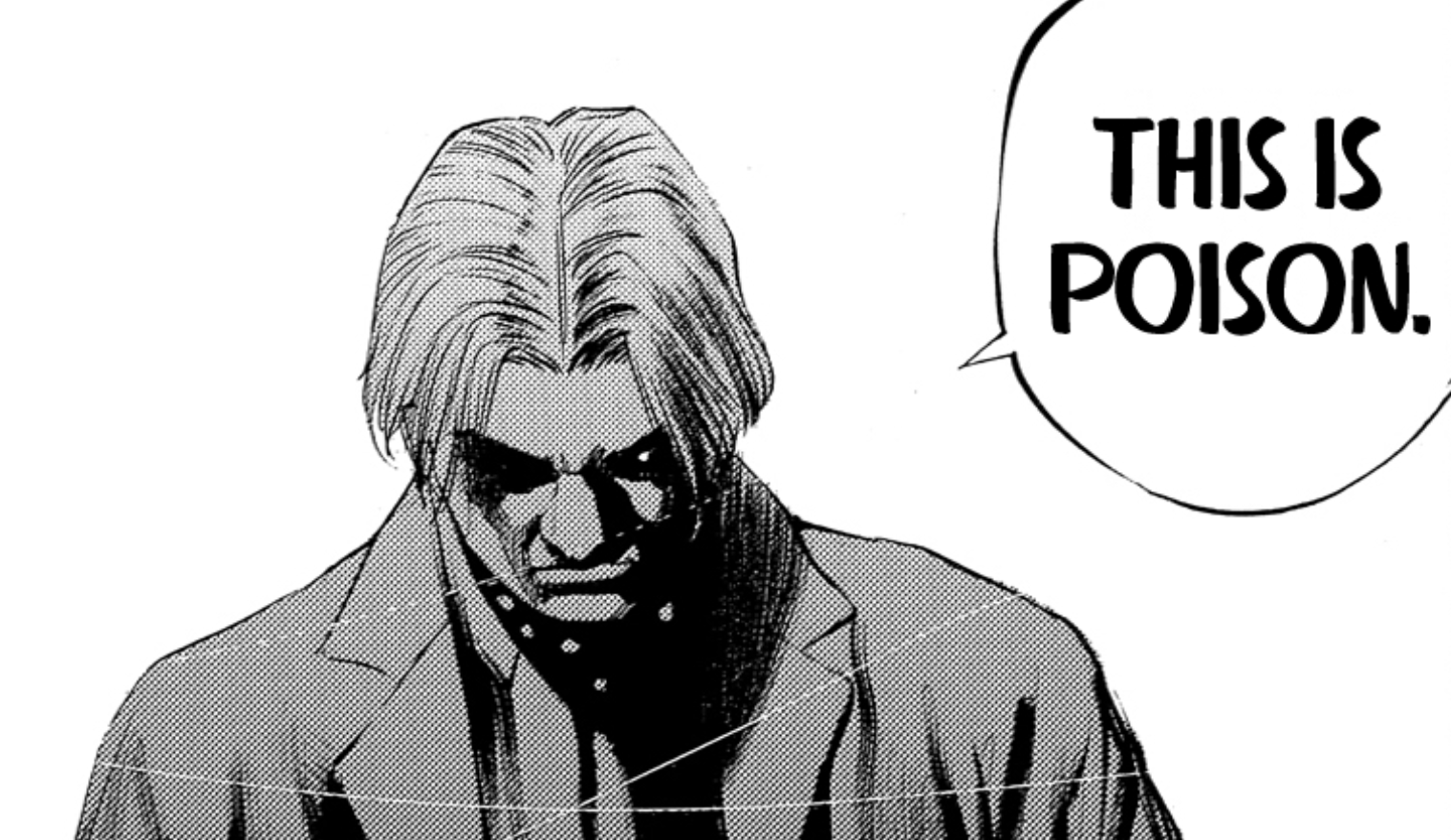}
  \end{minipage}
  
  \vspace{2mm}
  \hrule height 0.4pt
  \vspace{2mm}

  \noindent 
  \begin{minipage}[t][2.3cm][c]{0.33\linewidth}
    (E) \textit{\footnotesize Rumor has it that chocolate makes you 3x more productive when studying.}  \\
  \end{minipage}%
  \hfill
  \begin{minipage}[t][2.3cm][t]{0.3\linewidth}
    \textcolor{red}{\Large\ding{55}} \small{(0.2)} \\
    \includegraphics[width=\linewidth, height=0.8\textheight, keepaspectratio]{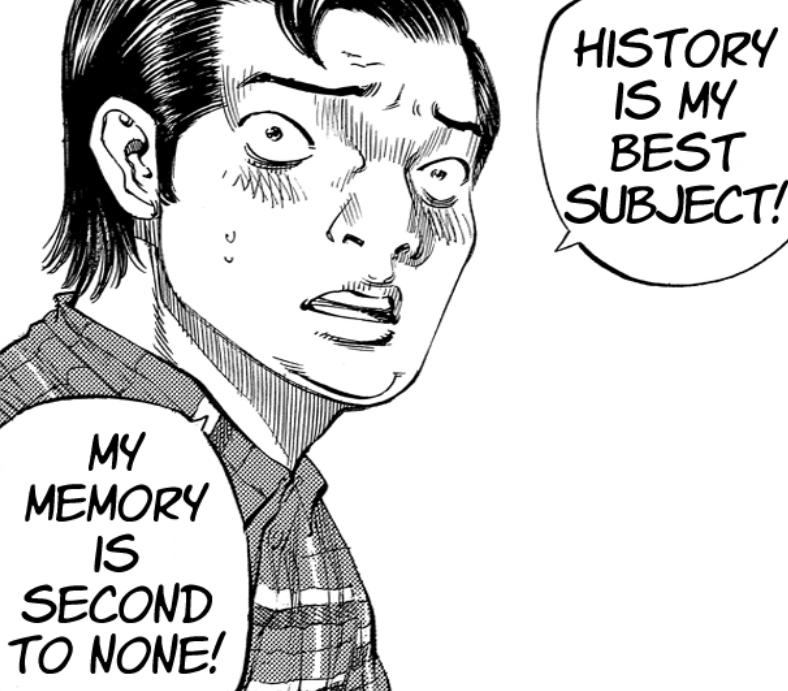}
  \end{minipage}%
  \hfill
  \begin{minipage}[t][2.3cm][t]{0.3\linewidth}
    \textcolor{green}{\Large\ding{51}} \small{(0.8)} \\
    \includegraphics[width=\linewidth, height=0.8\textheight, keepaspectratio]{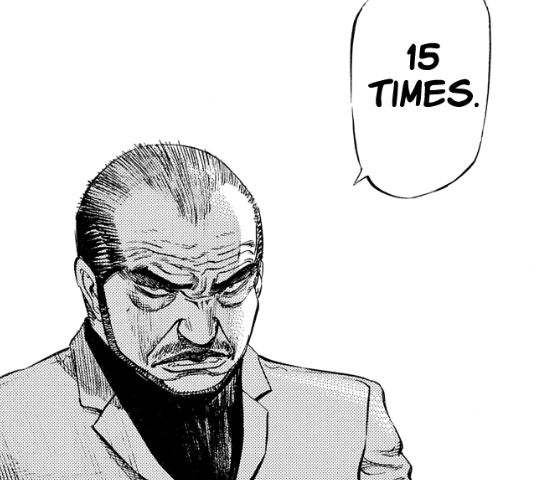}
  \end{minipage}
  
  \vspace{2mm}
  \hrule height 0.4pt
  \vspace{2mm}

  \noindent 
  \begin{minipage}[t][2.2cm][c]{0.33\linewidth}
    (F) \textit{\footnotesize Maybe life without a smartphone isn't so bad after all.}  \\
  \end{minipage}%
  \hfill
  \begin{minipage}[t][2.2cm][t]{0.3\linewidth}
    \textcolor{red}{\Large\ding{55}} \small{(0.2)} \\
    \includegraphics[width=\linewidth, height=0.8\textheight, keepaspectratio]{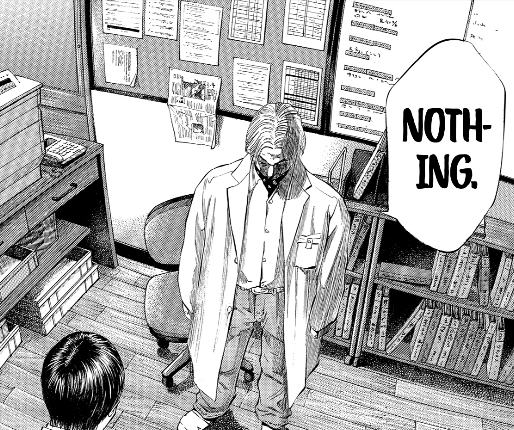}
  \end{minipage}%
  \hfill
  \begin{minipage}[t][2.2cm][t]{0.3\linewidth}
    \textcolor{green}{\Large\ding{51}} \small{(1.0)} \\
    \includegraphics[width=\linewidth, height=0.8\textheight, keepaspectratio]{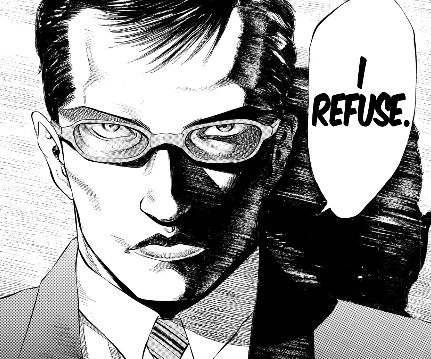}
  \end{minipage}

  \vspace{2mm}
  \hrule height 0.4pt
  \vspace{2mm}

    \noindent 
  \begin{minipage}[t][2cm][c]{0.33\linewidth}
    (G) \textit{\footnotesize Whoops, I gamed straight through to the morning.}  \\
  \end{minipage}%
  \hfill
  \begin{minipage}[t][2cm][t]{0.3\linewidth}
    \textcolor{red}{\Large\ding{55}} \small{(0.2)} \\
    \includegraphics[width=\linewidth, height=0.8\textheight, keepaspectratio]{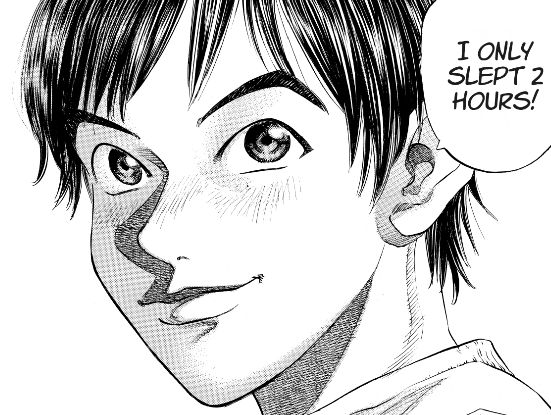}
  \end{minipage}%
  \hfill
  \begin{minipage}[t][2cm][t]{0.3\linewidth}
    \textcolor{red}{\Large\ding{55}} \small{(0.2)} \\
    \includegraphics[width=\linewidth, height=0.8\textheight, keepaspectratio]{figures/case_study/context-29_sim-pref-0.2_vol1-7p-en-resized.png}    
  \end{minipage}
  
  \vspace{1mm}
  \hrule height 0.6pt

  \caption{\small Examples of selected replies by Sarashina-Text-Emb (similarity-based) and GPT-5 (preference-based). Numbers in parentheses represent funniness score.}
  \label{tab:examples}
\end{table}

The \texttt{sim-select} model, Sarashina-Text-Emb, succeeds when its keyword-matching mechanism produces unexpectedly complex humor like personification or irony. In case (A), a simple match for ``\textit{fast}'' selects the meme ``\textit{HE'S JUST SO FAST...}'' The humor arose not from the word match itself, but because the reply personifies the Wi-Fi with an overly dramatic, human-like reaction. This pattern continued in case (B), where a more subtle match for ``\textit{lost}'' selects ``\textit{LATELY I DON'T UNDERSTAND MYSELF.}'' This reply unexpectedly reinterprets a simple problem as a deep philosophical one, producing humor through its sudden exaggeration. Finally, in case (C), a literal match for ``\textit{I give up}'' resulted in a blunt, agreeing reply (``\textit{GIVE UP.}''). The humor here comes from its cold, dismissive tone, which is the opposite of the supportive reaction expected from a typical conversational agent.

In contrast, the \texttt{pref-select} model GPT-5 excels at understanding context that goes beyond simple keywords. This contextual inference is clear in case (D). While the \texttt{sim-select} model failed due to a superficial connection between ``\textit{fridge}'' and ``\textit{cool},'' GPT-5 infers the dangerous nature of the ``\textit{10-year-old pudding}'' and provided the fitting reply, ``\textit{THIS IS POISON.}''. Another key strength of GPT-5 was its ability to create a humorous persona. This was illustrated in case (E), where the simple association made by Sarashina-Text-Emb between ``\textit{studying}'' and related keywords like ``\textit{memory},'' ``\textit{history},'' or ``\textit{subject}'' was not effective. GPT-5, however, adopts the persona of a confident expert making an overconfident claim. The core of the humor lies in its statement, ``\textit{15 TIMES.}''. It directly builds on the ``\textit{3x}'' in the context, absurdly escalating it with a larger, baseless number. This overconfidence makes the reply far funnier. Likewise, in case (F), the direct ``\textit{I REFUSE}'' to living without a smartphone created another strong, humorous personality. In these instances, the humor arose from the character the reply embodies, making it more effective than the simple associations of the \texttt{sim-select} model.

However, both models can fail when a contextually relevant reply is too obvious and lacks an element of surprise. This is evident in case (G), where for the context ``\textit{gamed straight through to the morning},'' both models had selected the most predictable meme: ``\textit{I ONLY SLEPT 2 HOURS...}''. While this reply is perfectly relevant, it is also highly predictable and therefore received a low humor score. This highlights a core challenge for automated humor generation: models need to distinguish simple contextual relevance from genuine, surprising humor.

In summary, these case studies show that the \texttt{sim-select} method provides reliable relevance but struggles to create high-impact humor consistently. The \texttt{pref-select} approach shows greater potential for more complex, context-aware humor, but it must also avoid predictable responses to be truly effective.

\section{Discussion}
\label{sec:discussion}
In this section, we discuss what our findings reveal about the broader challenges of modeling humor in online communication. We focus on two complementary perspectives: (1) what these results suggest about multimodal and social understanding in current systems, and (2) how they inform future directions for building more contextually aware conversational models.

\subsection{Implications for Computational Humor}
\label{sec:implications}
\paragraph{LLMs Show Preliminary Evidence of Capturing Social Cues}
Our results show that \texttt{pref-select} methods using LLMs demonstrate advantages over \texttt{sim-select} methods. While the difference in scores was small, our case studies suggest this advantage comes from a more advanced capability. LLMs appear to capture complex social cues such as irony and exaggeration, going beyond the surface-level semantic relevance that \texttt{sim-select} methods rely on. This finding suggests that computational models are moving beyond simple relevance matching and are starting to develop an ability to reason about the social and contextual nature of humor as it is used on the web.
These results suggest that LLMs can represent some social cues, but how these representations translate into explicit humor judgments remains an open question.

\paragraph{The Gap Between Understanding and Using Multimodal Memes}
Through our experiments, we identified the difficulty of using visual information. Providing visual information, either as text descriptions or direct images, failed to improve performance. It was particularly ineffective for models that process images directly. This points to a critical distinction between understanding and using visual humor. While prior work shows that models can achieve some success in interpretation tasks like classifying a meme's sentiment or explaining its humor~\cite{sharma-2023-what-do-you}, our findings suggest this understanding does not easily translate into effective action. The difficulty seems to arise when models must actively use socio-cultural cues from an image, such as exaggerated expressions or ironic situations, to distinguish the wittiest reply from a set of plausible candidates. Closing this gap between passive understanding and active contextual use remains a fundamental challenge as web communication becomes increasingly visual.

We hypothesize that three fundamental mismatches may explain why visual information fails to improve performance. First, there may be an \emph{objective mismatch}: models are trained on semantic similarity (matching objects and scenes), whereas humor requires pragmatic alignment through recontextualization, the act of placing fixed visual content in a new conversational context to create surprising juxtaposition. Second, there may be a \emph{data mismatch}: pre-training relies on descriptive captions that state explicit facts about images, while humor depends on implicit contextual cues that are not present in the image itself. Third, there may be a \emph{domain mismatch}: the uniform visual style of manga makes it difficult for general-purpose models to distinguish the fine-grained facial expressions and subtle visual cues that are essential for humor. Further investigation is needed to validate these hypotheses. We also note that these findings may be specific to our manga-based benchmark, and results could differ with other meme formats such as photographic memes or reaction GIFs.

Addressing this limitation may require new objectives or architectures that couple visual recognition with contextual inference, rather than simply scaling multimodal encoders.

\paragraph{From Simple Choices to Nuanced Judgments}
Finally, our experiments identify the core challenge for current models: distinguishing subtle differences in wit among semantically similar candidates. While the models demonstrated near-human performance in controlled situations with clearly distinguishable choices (Exp3), their ability to select the best option decreased in more realistic scenarios, such as the retrieve-and-rerank approach in Exp2 or when candidate similarity was high in Exp4. This suggests that the retrieve-and-rerank architecture itself is a promising approach. The initial retrieval is effective at narrowing the pool to relevant and often humorous content. The main difficulty, therefore, lies in the final selection step. Future work should focus on designing evaluation settings that better capture these subtle humor distinctions, enabling more reliable progress across models.

\subsection{Limitations and Future Directions}
\label{sec:limitations}
\paragraph{Scope of the Benchmark}
The design of our benchmark involves intentional scope limitations. First, we focused on a single domain: freely licensed Japanese manga panels. This choice was made to isolate humor signals in a controlled setting: unlike generative memes, manga panels have fixed content, which is ideal for isolating selection capabilities rather than generation. As subjectivity was high even in this controlled setting, adding cultural diversity prematurely would have made consensus impossible. This controlled framework can serve as a baseline for future cross-cultural studies. Online meme formats are diverse, and each creates humor through different conventions, such as image-macro captions or ongoing narratives. A key direction for future work is to expand this scope to other formats, like GIFs or reaction photos, to analyze how each generates its own unique humorous social dynamics. Second, we used a binary ``funny/not funny'' scale. This approach bases the subjective task of humor evaluation on a clearer, more observable reaction (whether the content elicits a laugh), which simplifies judgment for annotators and reduces subjective variance. Future benchmarks could use more detailed annotations to gain deeper insights. For example, graded scales could be used to analyze humor intensity, while the categorization of humor types could allow for a more detailed analysis of which humorous strategies are most effective in different contexts.

Ultimately, this research aims to enable systems that can produce 
genuinely funny replies in real-world interactions. While our use of synthetic 
contexts provides a controlled setting for this initial investigation, 
validating these findings in naturalistic environments remains an important 
next step.

\paragraph{Towards More Socially Aware Agents}
The practical implication of this research lies in improving how conversational systems respond to social and multimodal cues in online communication. A natural next step is to generate or select visual replies that align with contextual humor, extending beyond the simple retrieve-and-rank baseline explored in this study. To make such systems practical, several open challenges remain. These include modeling user and cultural context, integrating multimodal reasoning across text and images, and handling the temporal dynamics of memes, whose formats and popularity change rapidly over time. Further progress will also depend on developing evaluation frameworks that reflect these evolving conditions rather than static benchmarks alone. Beyond these technical challenges, practical deployment also requires attention to potential risks, such as selecting inappropriate or offensive content in sensitive contexts; human oversight and context-aware filtering will be essential.
The \textsc{MaMe-Re} benchmark provides a foundation for addressing these issues under controlled conditions, serving as a bridge between computational humor research and socially aware dialogue systems.

\section{Conclusion}
\label{sec:conclusion}
Humor is a defining element of human communication, and our work takes an early step toward enabling machines to recognize and participate in this social form of expression.
In this work, we proposed a new research direction for computational meme analysis, shifting the focus from static content to dynamic, interactive communication. To support this direction, we introduced the Meme Reply Selection task and the \textsc{MaMe-Re} benchmark. This work establishes a systematic framework for studying contextual humor in online conversations. Our analysis revealed several fundamental challenges for current models. These include the gap between passively understanding and actively using visual humor, and the difficulty of distinguishing subtle differences in wit among similar candidates. We offer \textsc{MaMe-Re} as a benchmark to measure progress on these fundamental challenges and to advance research on how AI systems understand and participate in humor within human communication.

\bibliography{AnonymousSubmission/Latex/aaai2026}

\end{document}